\documentclass[lettersize,journal]{IEEEtran}
\usepackage{amsmath,amsfonts}
\usepackage{array}
\usepackage[caption=false,font=normalsize,labelfont=sf,textfont=sf]{subfig}
\usepackage{subfig}
\usepackage{textcomp}
\usepackage{stfloats}
\usepackage{url}
\usepackage{verbatim}
\usepackage{graphicx}
\usepackage{cite}
\usepackage{bbding}
\usepackage{booktabs}
\usepackage{color}
\usepackage{xcolor}
\usepackage{colortbl}
\usepackage{amsfonts}
\usepackage{amsmath}
\usepackage[ruled,linesnumbered]{algorithm2e}
\usepackage{newfloat}
\usepackage{listings}
\usepackage[colorlinks,urlcolor=blue,linkcolor=blue,citecolor=blue]{hyperref}
\usepackage{orcidlink}

\newcommand{\revise}[1]{{\color{black}#1}}

\begin{document}

\title{Progressive Conditioned Scale-Shift Recalibration of Self-Attention for Online Test-time Adaptation}

\author{Yushun Tang$^{\orcidlink{0000-0002-8350-7637}}$, Ziqiong Liu$^{\orcidlink{0009-0002-0418-4858}}$, Jiyuan Jia$^{\orcidlink{}}$, Yi Zhang$^{\orcidlink{0000-0002-5831-0170}}$, and Zhihai He$^{\orcidlink{0000-0002-2647-8286}}$,~\IEEEmembership{Fellow, IEEE}
\thanks{This work was supported by the National Natural Science Foundation of China (No. 62331014) and Project 2021JC02X103. \textit{(Corresponding author: Zhihai He.)}}

\thanks{
Yushun Tang, Ziqiong Liu, Jiyuan Jia, and Yi Zhang are with the Department of Electrical and Electronic Engineering, Southern University of Science and Technology, Shenzhen 518055, China.}

\thanks{Zhihai He is with the Department of Electrical and Electronic Engineering, Southern University of Science and Technology, Shenzhen 518055, China, and also with Pengcheng Lab, Shenzhen 518066, China (e-mail: hezh@sustech.edu.cn).}



}

\markboth{IEEE Transactions on Multimedia}%
{Shell \MakeLowercase{\textit{et al.}}: A Sample Article Using IEEEtran.cls for IEEE Journals}


\maketitle

\begin{abstract}
Online test-time adaptation aims to dynamically adjust a network model in real-time based on sequential input samples during the inference stage. In this work, we find that, when applying a transformer network model to a new target domain, the Query, Key, and Value features of its self-attention module often change significantly from those in the source domain, leading to substantial performance degradation of the transformer model.  To address this important issue, we propose to develop a new approach to progressively recalibrate the self-attention at each layer using a local linear transform parameterized by conditioned scale and shift factors. We consider the online model adaptation from the source domain to the target domain as a progressive domain shift separation process. At each transformer network layer, we learn a Domain Separation Network to extract the domain shift feature, which is used to predict the scale and shift parameters for self-attention recalibration using a Factor Generator Network. These two lightweight networks are adapted online during inference. Experimental results on benchmark datasets demonstrate that the proposed progressive conditioned scale-shift recalibration (PCSR) method is able to significantly improve the online test-time domain adaptation performance by a large margin of up to 3.9\% in classification accuracy on the ImageNet-C dataset. 
\end{abstract}


\section{Introduction}
\label{sec:intro}
Domain adaptation is an important and challenging research task in machine learning and computer vision. Domain shifts often result in significant performance declines when the target domain, where the model is tested, substantially diverges from the source domain, where it was initially trained \cite{mirza2021robustness,zhang2021adaptive}. Existing methods have been largely focused on unsupervised domain adaptation \cite{liang2020we,tang2023cross,wang2022cross,active_zhou,guided_zhang}. Online test-time adaptation is an emerging machine-learning topic that dynamically adjusts a network model in real-time during the inference stage \cite{wang2021tent,tang2023neuro,wen2023test,tang2024learning,meng2022dual,tang2024domain,tang2025dual}. 

Online test-time adaptation has been explored in various machine learning tasks and application scenarios \cite{wang2021tent,shin2022mm,liang2024comprehensive}. \revise{The TENT \cite{wang2021tent} method updates the layer normalization module by minimizing entropy loss}, while the MEMO method \cite{zhang2022memo} optimizes the entropy of averaged predictions over multiple random augmentations of input samples. The VMP method \cite{jing2022variational} introduces perturbations into model parameters using variational Bayesian inference. Additionally, the SAR method \cite{niutowards} proposes a sharpness-aware and reliable optimization scheme, which discards samples with significant gradients and encourages model weights to converge to a flat minimum. Most of these existing methods focus on ResNet-like network models. 

\begin{figure}[!t]
    \centering
    \includegraphics[width=\linewidth]{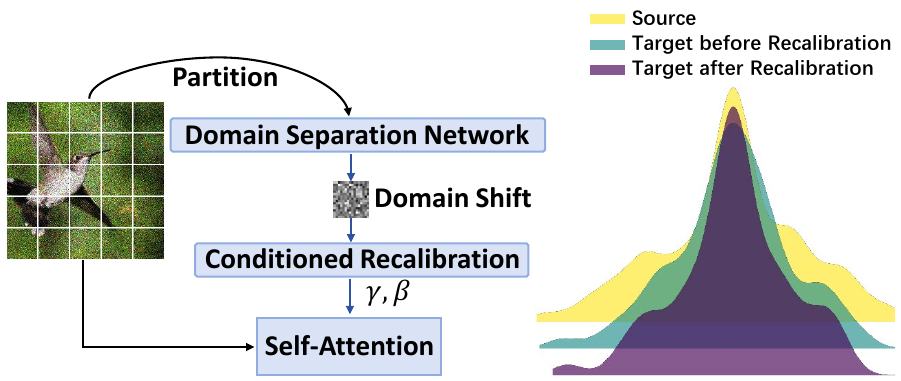}
    \caption{An illustration of our proposed Progressive Conditioned Scale-Shift Recalibration of Self-Attention. An example of a Query vector distribution in the target ImageNet-C dataset shows significant deviation from the source ImageNet domain. However, after applying our conditioned scale-shift recalibration, the distribution aligns more closely with the source domain, leading to improved generalization performance.}
    \label{fig:intro}
\end{figure}

Transformer-based methods have recently achieved remarkable success in various machine learning tasks, largely owing to their powerful self-attention mechanism. This mechanism allows transformers to capture long-range dependencies and contextual relationships within data, making them highly effective for tasks such as natural language processing \cite{vaswani2017attention}. Vision Transformers (ViTs), which apply the transformer architecture to image data, have  demonstrated superior performance on a wide range of tasks, such as  image classification, object detection, and segmentation, outperforming traditional convolutional neural networks (CNNs) \cite{dosovitskiyimage}. Despite these successes, the generalization ability of transformer models across different domains remains a significant challenge. When a transformer model trained on the source domain is applied to a new unseen target domain, the performance often degrades substantially. This issue arises because the Query, Key, and Value features within the self-attention mechanism, which are crucial for capturing the relationships between different parts of the input data, are highly sensitive to the specific characteristics of the training data. In the target domain, where these characteristics  change, the Query, Key, and Value features can diverge significantly from those in the source domain, leading to a mismatch in the learned representations. For example, a Vision Transformer trained on clean images in the source domain may struggle when deployed in a target domain with noisy, low-resolution images, as the Query, Key, and Value features extracted in such scenarios might not align with those learned during training. This problem becomes even more pronounced in real-world applications, where domain shifts are common and can be unpredictable. Addressing this critical challenge in domain adaptation for transformer networks necessitates innovative solutions to recalibrate features and self-attention during inference. This raises an important research question:  \textit{How can we efficiently recalibrate these features and self-attention mechanisms, perturbed by domain shifts, to enhance the online test-time adaptation performance of transformer models?}


To address this challenge, we propose a novel approach called \textit{Progressive Conditioned  Scale-Shift Recalibration} for online test-time adaptation of transformer models. As shown in Figure \ref{fig:intro}, this method aims to dynamically adjust the self-attention mechanisms within transformers as they process new input data from different domains. The core idea is to recalibrate the Query, Key, and Value components of the self-attention module, which are crucial for capturing contextual relationships in the data but are highly sensitive to domain shifts.
To counter this, we introduce a scale and a shift factor into the Query, Key, and Value components of the self-attention module. These components determine how much attention each part of the input sequence should receive relative to others, and any misalignment caused by domain shifts can significantly degrade model performance. Specifically, the scale factors adjust the magnitude of the Query, Key, and Value features, ensuring that their influence is appropriately balanced, while the shift factors modify their shift values, correcting for any systematic biases introduced by the domain shift.
A key challenge is how to derive optimal scale and shift factors. To achieve this, we draw inspiration from recent advances by \textit{adaLN} in the generative model \cite{brock2018large,karras2019style,peebles2023scalable} and Conditional Context Optimization in the vision-language model \cite{zhou2022conditional}, we develop a Factor Generator Network that regresses these factors conditioned on the domain-specific feature. Additionally, drawing from Fermat-Webber point denoising in signal filtering \cite{lipman2007parameterization,kim2007study,kwon2008regularization}, we introduce a \revise{Domain Separation Network} to generate a Domain token derived from all the patch tokens at each transformer network layer. Once the Domain token is obtained, we train the Factor Generator Network to produce the scale and shift factors conditioned on both the Domain token and the Class token. Our findings indicate that these scale and shift factors, applied at each transformer network layer, gradually mitigate the impact of domain shift during online test-time adaptation, largely restoring the original features. This gradual adjustment process enables the recovery of the original self-attention, allowing the model to maintain performance across diverse domains. Extensive experimental results demonstrate that the proposed  Progressive Conditioned Scale-Shift Recalibration (PCSR) method significantly enhances test-time domain adaptation performance, outperforming existing state-of-the-art methods by a large margin.

\section{Related work and Major Contributions}
\label{sec:related_work}
This work is related to test-time adaptation, source-free unsupervised domain adaptation, and parameter-efficient fine-tuning. 

\subsection{Test-time Adaptation}
Models trained on source domains often experience significant performance degradation when deployed to target domains with different distributions \cite{ben2006analysis}. To address this issue, Test-time training (TTT) and Test-time adaptation (TTA) methods have been introduced, which dynamically adjust the model during testing to handle distribution shifts. TTT was first introduced in \cite{sun2020test}, where the model is trained during the testing phase using a self-supervised task to update the network parameters of the feature extractor, thereby mitigating the effects of distribution shifts. TTT++ \cite{liu2021ttt++} introduced a test-time feature alignment strategy and contrastive learning, which effectively alleviated the performance degradation problem when facing severe distribution changes. TTA-MAE \cite{gandelsman2022test} implemented self-supervised learning through a masked autoencoder based on the Transformer structure.
Unlike the above TTT methods that rely on source domain information, TTA adapts using only target domain data. \revise{Tent \cite{wang2021tent} first proposed TTA and achieves model adaptation by adjusting layer normalization (LN) parameters during testing to minimize the entropy of the model's predictions}, thereby accommodating different data distributions. CTTDA \cite{wang2022continual} introduces a method that continuously adjusts batch normalization parameters and leverages diverse self-supervised learning tasks during test time to address ongoing distribution shifts. 
To address performance drops due to imbalanced samples and distribution shifts, SAR \cite{niu2023towards} removes high-gradient noise samples and guides model weights toward smoother minima. The VCT method \cite{tang2024learning} updates the Class Token during the testing stage with long and short-term optimization. Applied to pre-trained vision-language models, \cite{karmanov2024efficient} proposed the Test-time Dynamic Adapter (TDA), which leverages a lightweight key-value cache system for progressive pseudo label optimization and uses a negative pseudo labeling mechanism to handle label noise. In this work, unlike existing methods that adapt the pre-trained normalization layers, we introduce new conditioned scale and shift factors to the Query, Key, and Value components of the self-attention module, guided by reliable entropy loss.

\subsection{Source-free Unsupervised Domain Adaptation}
Source-free unsupervised domain adaptation (source-free UDA) aims to adapt the model trained on the source domain to the unlabeled target domains without leveraging the source data \cite{liang2020we,sun2022prior,Yang_2021_ICCV,li2020model,huang2021model}. 
The SHOT method~\cite{liang2020we} computed pseudo labels through the nearest centroid classifier and optimized the model with information maximization criteria.
The KUDA method \cite{sun2022prior} utilized the prior knowledge about label distribution to refine model-generated pseudo labels.
The SFDA-DE method~\cite{ding2022source} aligned domains by estimating source class-conditioned feature distribution. 
The HCL method~\cite{huang2021model} proposed a solution for addressing the lack of source data by introducing both instance-level and category-level historical contrastive learning. The DIPE method~\cite{Wang_2022_CVPR} focuses on exploring the domain-invariant parameters of the model, rather than trying to learn domain-invariant representations. The CRS method \cite{Zhang_2023_CVPR} transfers the domain-invariant class relationship by class-aware and instance discrimination contrastive learning. The method proposed by \cite{Litrico_2023_CVPR} refines pseudo-labels by loss reweighting strategy.
These source-free methods are offline, requiring access to the complete test dataset. It also costs a number of epochs for model adaptation. 
All of the source-free UDA methods first train the source model in the source domain and then adapt the model in the target domain with multiple epochs. 
In contrast, our fully online test time adaptation adapts the given source model on the fly during testing which only accesses the test samples once.

\subsection{Parameter-Efficient Fine-tuning}
Parameter-efficient fine-tuning (PEFT) has emerged as a significant area of research in deep learning, particularly in scenarios where models are deployed on resource-constrained devices or when training needs to be performed on large-scale models with limited computational resources \cite{houlsby2019parameter,hu2022lora,xin2024parameter}. The primary goal of PEFT is to adapt pre-trained models to new tasks or domains with minimal changes to the model's parameters, thereby reducing the computational and memory overhead associated with fine-tuning.
These methods typically freeze most of the original model's parameters, selectively updating a small subset of parameters or introducing a few additional parameters for new tasks, thereby ensuring both efficiency and effectiveness of the model. For example, Low-rank Adaptation (LoRA) \cite{hu2022lora} reduces the parameter update burden by decomposing the model's weight matrices into products of low-rank matrices, thus lowering the computational load.  The Adapter method, introduced by \cite{houlsby2019parameter}, achieves fine-tuning by inserting lightweight adapter modules into specific layers of the model. This approach adds only a small number of trainable parameters for each task while keeping the rest of the model's parameters fixed, thereby enabling efficient parameter sharing across different tasks. The AdaptFormer \cite{chen2022adaptformer} replaces the original MLP block with an AdaptMLP module, which includes a trainable down-up bottleneck structure, enabling efficient fine-tuning of ViT models for a variety of visual tasks. DAPT \cite{zhou2024dynamic} combines Dynamic Adapter and prompt tuning methods in point cloud analysis to achieve parameter-efficient transfer learning.  Some methods inject learnable parameters into the pre-trained model or the input image, providing an alternative fine-tuning way. For instance, Visual Prompt Learning (VPT) \cite{jia2022visual,gao2022visual,10204082,sun2023vpa} incorporates task-specific learnable parameters directly into the input sequence at each layer of the Vision Transformer (ViT) encoder.
Additionally, some methods do not introduce new modules but instead utilize sparsification to identify and fine-tune the parameters that are most crucial to the model. Diff Pruning \cite{guo2020parameter} achieves this by learning a sparse difference mask (diff mask) to selectively adjust the necessary parameters, while BitFit \cite{zaken2022bitfit} performs sparse fine-tuning by modifying only the bias terms.
All of these parameter-efficient fine-tuning methods adapted pre-trained models to downstream tasks through supervised learning. In this work, we perform a parameter-efficient online test-time adaptation in an unsupervised manner.


\subsection{Major Contributions}
In this work, we propose an approach for online test-time adaptation of transformer models by designing adaptable scale and shift factors for self-attention modules. Compared to existing methods, the major contributions of this work are as follows: (1) We introduce a lightweight neural network called the Domain Separation Network, which generates a Domain token representing the domain shift, derived from all image patches at each layer. This Domain token effectively captures domain-specific features, facilitating the learning of improved scale and shift factors. (2) We develop a Factor Generator Network that produces these recalibration factors conditioned on both the Domain token and the Class token. (3) We present a novel self-attention module design for transformer networks, which recalibrates the Query, Key, and Value features of test samples online during inference in the target domain. (4) Our experimental results demonstrate that the proposed Progressive Conditioned Scale-Shift Recalibration method significantly enhances the domain adaptation performance of transformer models, outperforming state-of-the-art methods in online test-time domain adaptation across multiple popular benchmark datasets.

\section{Method}
In this section, we present our method of Progressive Conditioned Scale-Shift Recalibration for online test-time adaptation in detail.

\begin{figure*}[!ht]
    \centering
    \includegraphics[width=\linewidth]{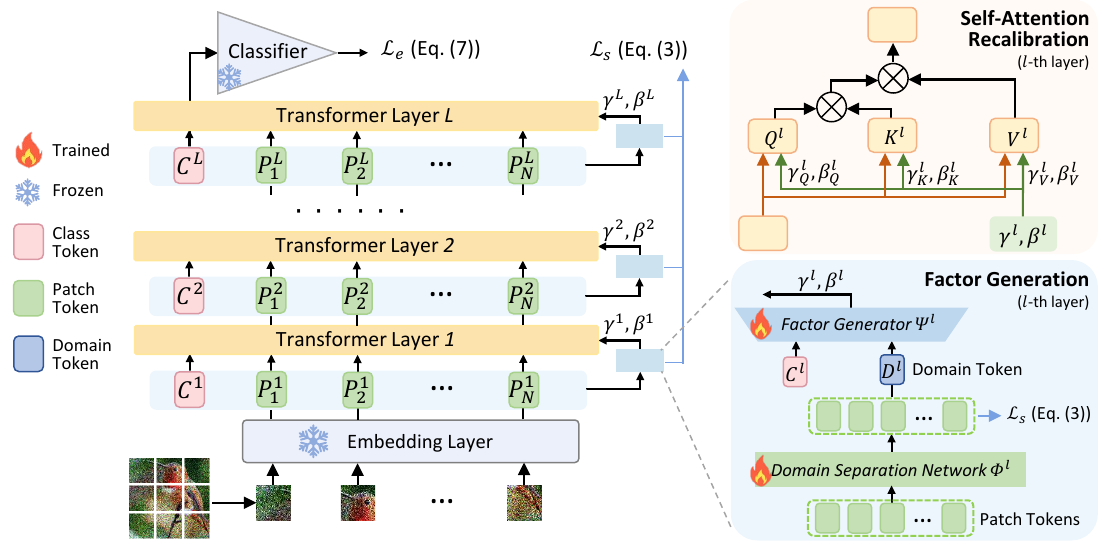}
    \caption{An overview of the proposed Progressive Conditioned Scale-Shift Recalibration method. We introduce a Factor Generation Block including the Domain Separation Network $\Phi_l$ and the Factor Generator Network $\Psi_l$ for each transformer layer. The Domain Separation Network aims to generate the Domain token, representing the domain-specific features. The Factor Generator Network aims to generate the scale and shift factors for the Query, Key, and Value components. These two networks are updated online during inference, guided by the reliable entropy loss and the similarity loss.}
    \label{fig:overview}
\end{figure*}

\subsection{Method Overview}
Suppose we have a model $\mathcal{M} = f_{\theta_s}(y|X_s)$ with parameters $\theta_s$, which has been successfully trained on source data $\{X_s\}$ with corresponding labels $\{Y_s\}$. During fully online test-time adaptation, we are provided with target data $\{X_t\}$ along with unknown labels $\{Y_t\}$. Our objective is to adapt the trained model in an unsupervised manner during testing.
In this scenario, we receive a sequence of input sample batches $\{\mathbf{B}_1, \mathbf{B}_2, ..., \mathbf{B}_T\}$ from the target data $\{X_t\}$. It should be noted that during each adaptation step $t$, the network model can only rely on the $t$-th batch of test samples, denoted as $\mathbf{B}_t$. 

Our method aims to dynamically adjust the Vision Transformer model during inference to mitigate performance degradation caused by domain shifts. The key innovation in our method lies in its ability to recalibrate the self-attention mechanism of the ViT in response to these shifts, ensuring that the model maintains high performance even when applied to new or unseen domains. As shown in Figure \ref{fig:overview}, this recalibration is achieved through a series of steps that involve generating a domain shift representation, conditioning a network to produce adjustment factors, and applying these factors to the model's self-attention components. The central challenge lies in learning the domain-specific shift feature to recalibrate the Query, Key, and Value features. To address this, we first develop a Domain Separation Network to generate the Domain token, representing the domain shift derived from all patch tokens at each transformer network layer, as all patches within an image share the same domain shift. Once the Domain token is obtained, we train a Factor Generator Network to produce the scale and shift factors conditioned on the Domain token and the Class token. By applying these scale and shift factors to the self-attention components, the feature domain shift at each transformer layer is gradually mitigated. In the following sections, we will introduce our method in detail.

\subsection{Learning to Separate Domain Shift}
\begin{figure}
    \centering
    \includegraphics[width=\linewidth]{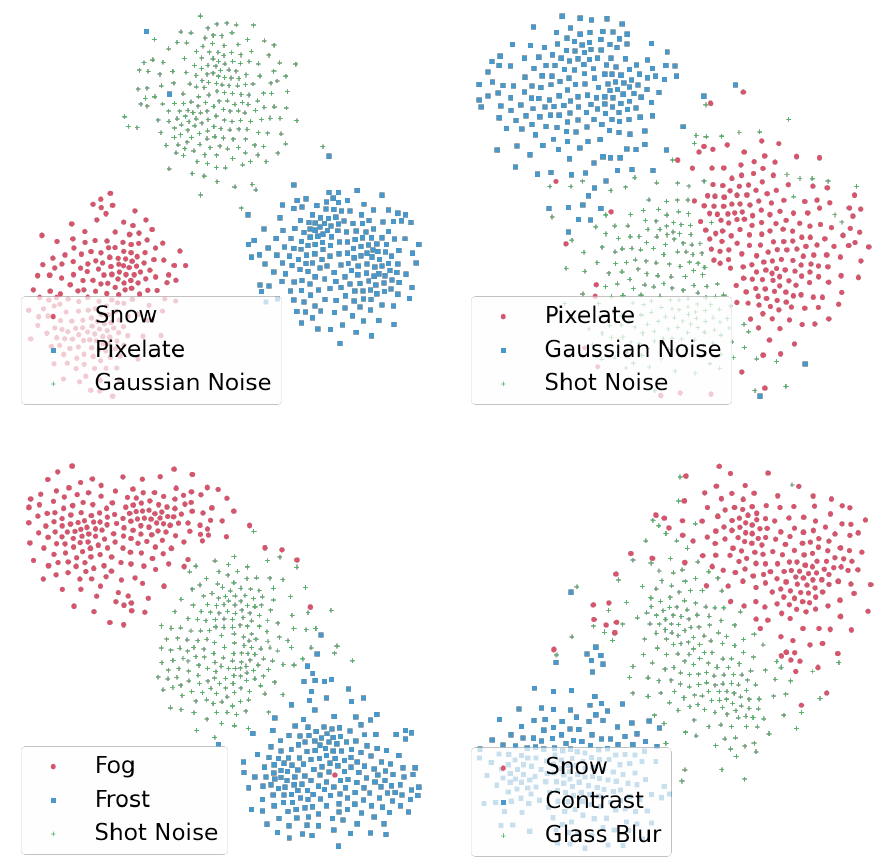}
    \caption{T-SNE visualization of the Domain token in the adaptation process for different domains in ImageNet-C dataset. Each corruption represents a domain.}
    \label{fig:domain_token}
\end{figure}

We train the Domain Separation Network to obtain the Domain token inspired by the Fermat-Weber point location problem.
This method has been utilized to distill a representative summary from a large set of samples in the presence of noise and outliers \cite{lipman2007parameterization,kim2007study,kwon2008regularization}. In the context of Vision Transformers (ViTs), this principle can be applied to extract a representative feature from a set of patch tokens, each of which is affected by the same domain shift or noise. This representative feature, which we term the Domain token, encapsulates the domain-specific information in each transformer layer. Our goal is to distill a representative feature from these patches that captures the domain shift. In this work, we introduce a Domain Separation Network $\Phi^l$ to generate the Domain token, representing the domain-specific information in each transformer layer $l$. Specifically, the Domain token is denoted as:
\begin{equation}
    D^l = Avg[\Phi^l(P_1^l, P_2^l, \cdots, P_N^l)],
\end{equation}
where $P_n^l$ represents the $n$-th patch feature in the $l$-th layer of transformer, and $Avg[.]$ represents average pooling. By aggregating these patch tokens, the Domain Separation Network produces a summary token that effectively captures the overall domain shift affecting the input image at that particular layer. This process allows us to condense the information from multiple patches, which all share the same domain conditions, into a single, robust representation.
Similar to the Fermat-Weber point location problem, we use cosine distance to measure the distance between points in the patch feature space. We aim for the learned Domain token to have the minimum distance to each patch feature point. To achieve this, we compute the cosine similarity matrix $M$ for each pair of features after the Domain Separation Network. The cosine similarity between two patch token domain features $f_j^l$ and $f_k^l$ at layer $l$ is given by:
\begin{equation}
    M^{l}_{jk} = \frac{f_j^l \cdot f_k^l}{\parallel f_j^l \parallel \cdot \parallel f_k^l \parallel},
\end{equation}
where $\cdot$ denotes the dot product, and $\parallel \cdot \parallel$ denotes the vector norm. The similarity matrix $M$ thus provides a pairwise comparison of all patch tokens within a layer, reflecting how closely related they are in the feature space.
To ensure that the Domain token effectively summarizes the domain shift, for a pre-trained ViT network model with $L$ layers, we introduce a similarity loss function to constraint the Domain Separation Networks as follows:
\begin{equation}
    \mathcal{L}_{s}(\theta_t; x) = -\frac{1}{L}\sum\limits_{l=1}^{L} \frac{1}{N^2} \sum\limits_{j=1}^{N} \sum\limits_{k=1}^{N} M^{l}_{jk},
\end{equation}
where $N$ is the patch number, and $M^{l}_{jk}$ represents the cosine similarity between $j$-th and $k$-th patch token features after the Domain Separation Network in the $l$-th layer. By minimizing $\mathcal{L}_{s}(\theta_t; x)$, the Domain Separation Network learns to produce a token that is centrally located in the feature space, making it an optimal summary of the domain shift. Figure \ref{fig:domain_token} presents the t-SNE plot visualization of the Domain token in the last layer, illustrating samples from various target domains with different domain corruptions, each represented by a different color. The plot reveals that samples from the same domain, despite belonging to entirely different classes, tend to cluster together. This observation indicates that the Domain token generated by the Domain Separation Network $\Phi^l$ effectively captures the domain-specific characteristics.
In the following section, we utilize the learned domain-specific Domain token to generate the scale and shift factors.

\subsection{Predicting the Scale-Shift Factors}
The Factor Generator Network is an essential component of our proposed Progressive Conditioned Scale-Shift Recalibration method, enabling the Vision Transformer to dynamically adjust to domain shifts during inference. The primary function of this network is to generate the scale factors $\gamma$ and shift factors $\beta$ that are used to recalibrate the Query, Key, and Value components of the self-attention module within each layer of the transformer. These adjustments help mitigate the impact of domain shifts, thereby enhancing the model's performance across different domains. Once the Domain token is learned, we train another Factor Generator Network to generate these scale and shift factors conditioned by the Domain token and the Class token, as the Class token also contains domain knowledge.  The Factor Generator Network is a lightweight fully connected layer, denoted as $\Psi^l$, which takes the Domain token $D^l$ and the Class token $C^l$ as inputs and outputs the scale factors $\gamma^l$ and shift factors $\beta^l$ in the $l$-th layer. Mathematically, this process is described by the following equation:
\begin{equation}
    [\gamma^l, \beta^l] = \Psi^l(D^l, C^l).
\end{equation}
The scale factors and shift factors are then applied to the Query, Key, and Value components of the self-attention module. Specifically, the Factor Generator Network $\Psi^l$ for each layer is a lightweight fully connected layer with input dimension $d$ and output dimension $2d$.

\subsection{Scale-Shift Recalibration of the Self-Attention Module}
The self-attention mechanism is central to the performance of transformer models. It relies on the Query, Key, and Value components to compute attention weights and generate contextualized representations. However, these components are highly sensitive to domain shifts, leading to degraded performance in novel domains.
To address the sensitivity of the Query, Key, and Value components to domain shifts, we introduce scale and shift factors into the self-attention module. These factors adjust the scale and shift of the Query, Key, and Value components, aligning them more closely with the source domain's characteristics. Formally, for a given input feature, the recalibrated Query, Key, and Value components are computed as:
\begin{equation}
\begin{cases}
    Q' = \gamma_q \cdot Q + \beta_q,\\
    K' = \gamma_k \cdot K + \beta_k,\\
    V' = \gamma_v \cdot V + \beta_v,\\
\end{cases}
\end{equation}
where $\gamma_q, \gamma_k, \gamma_v$ are the scale factors and $\beta_q, \beta_k, \beta_v$ are the shift factors for the Query, Key, and Value components, respectively. In this work, these factors are shared among the Query, Key, and Value components to address the common domain shift. Then we perform the recalibrated self-attention:
\begin{equation}
    \text{Attention}(Q', K', V') = \text{Softmax}\left(\frac{Q'K'^\top}{\sqrt{d}}\right)V',
\end{equation}
where $d$ represents the dimensions of the query, key, and value.
During the testing stage, we select the widely used entropy loss in other TTA methods to guide the adaptation \cite{niu2023towards}. For samples $x$ in the current mini-batch $\mathbf{B}_t$, the reliable entropy loss is defined as:
\begin{equation}\label{eq: reliable-loss}
     \mathcal{L}_e(\theta_t; x) = \mathbf{I}\cdot \mathbb{E}(\theta_t; x),
\end{equation}
where $\mathbf{I} = \mathbb{I}[\mathbb{E}(\theta_t; x)<E_0]$ is the mask to filter out test samples whose entropy is larger than the threshold $E_0$, $\mathbb{E}$ is the entropy function. 
On top of the entropy loss $\mathcal{L}_e(\theta_t; x)$, the combination of the similarity loss $\mathcal{L}_{s}(\theta_t; x)$ is used to encourage the Domain Separation Network to learn the domain-specific information, which leads to the following loss function:
\begin{equation}\label{eq:loss}
    \mathcal{L}(\theta_t; x) =  \mathcal{L}_{e}(\theta_t; x) + \lambda \cdot \mathcal{L}_{s}(\theta_t; x),
\end{equation}
where $\lambda = \frac{\sum \mathbf{I}}{B}$ serves as the balance parameter, with $\sum \mathbf{I}$ representing the sum of the entropy masks across the batch.
By incorporating the scale and shift factors at each transformer layer, our method gradually mitigates the impact of domain shifts. This incremental adjustment enables the model to recover the original feature representations, thereby maintaining performance across diverse domains.

\section{Experiments}
\label{sec:exp}
In this section, we conduct experiments on multiple online test-time adaptation settings and multiple dataset benchmarks to evaluate the performance of our proposed method.

\subsection{Benchmark Datasets and Baselines} 
In our experiments, following other TTA methods, we use the widely adopted \textbf{ImageNet-C} \cite{hendrycks2018benchmarking} benchmark dataset that covers all categories in ImageNet-1k, featuring fifteen types of degradation, five levels of degradation intensity, and 50,000 images per level of degradation. Additionally, we test on 30,000 images from \textbf{ImageNet-R} \cite{hendrycks2021many}, a dataset that explores complex and diverse visual styles, thereby presenting greater challenges in image content judgment. Another dataset we employ is \textbf{ImageNet-A} \cite{hendrycks2021natural}, consisting of 30,000 highly challenging images selected from ImageNet-1k. Furthermore, we conduct experiments on the \textbf{VisDA-2021} \cite{bashkirova2022visda} dataset, which includes 20,000 images and focuses on evaluating domain adaptation performance. \revise{In our experiments, we primarily compare our proposed method with the no adaptation source model, T3A \cite{iwasawa2021test}, CoTTA  \cite{wang2022continual}, DDA \cite{gao2022back}, MEMO \cite{zhang2022memo}, TENT \cite{wang2021tent}, Ada-Contrast \cite{chen2022contrastive}, CFA \cite{kojima2022robustvit}, DePT-G \cite{gao2022visual}, and SAR \cite{niutowards}.} Except for the baseline, all methods are online TTA methods.

\begin{table*}[!htbp]
\begin{center}
\caption{Classification Accuracy (\%) for each corruption in \textbf{ImageNet-C} at the highest severity (Level 5). The best result is shown in \textbf{bold}.}
\label{table: normal}
\resizebox{\linewidth}{!}
{
\begin{tabular}{l|ccccccccccccccc|c}
\toprule
Method & gaus & shot & impul & defcs & gls & mtn & zm & snw & frst & fg & brt & cnt & els & px & jpg & Avg.\\	
\midrule
Source & 46.9 & 47.6 & 46.9 & 42.7 & 34.2 & 50.5 & 44.7 & 56.9 & 52.6 & 56.5 & 76.1 & 31.8 & 46.7 & 65.5 & 66.0 & 51.0 \\
T3A & 16.6 & 11.8 & 16.4 & 29.9 & 24.3 & 34.5 & 28.5 & 15.9 & 27.0 & 49.1 & 56.1 & 44.8 & 33.3 & 45.1 & 49.4 & 32.2\\
CoTTA & 40.3 & 31.8 & 39.6 & 35.5 & 33.1 & 46.9 & 37.3 & 2.9 & 46.4 & 59.1 & 71.7 & 55.5 & 46.4 & 59.4 & 59.0 & 44.4\\
DDA &   52.5  & 54.0  & 52.1  & 33.8  & 40.6  & 33.3  & 30.2  & 29.7  & 35.0  & 5.0  & 48.6  & 2.7  & 50.0  & 60.0  & 58.8  & 39.1 \\
MEMO & 58.1 & 59.1 & 58.5 	& 51.6 & 41.2 & 57.1 & 52.4 & 64.1 & 59.0 & 62.7 & \textbf{80.3} & 44.6 & 52.8 & 72.2 & 72.1 & 59.1 \\
AdaContrast & 54.4 & 55.8 & 55.8 & 52.5 & 42.2 & 58.7 & 54.3 & 64.6 & 60.1 & 66.4 & 76.8 & 53.7 & 61.7 & 71.9 & 69.6 & 59.9 \\
CFA  & 56.9 & 58.0 & 58.1 & 54.4 & 48.9 & 59.9 & 56.6 & 66.4 & 64.1 & 67.7 & 79.0 & 58.8 & 64.3 & 71.7 & 70.2 & 62.4 \\
TENT & 57.6 & 58.9 & 58.9 & 57.6 & 54.3 & 61.0 & 57.5 & 65.7 & 54.1 & 69.1 & 78.7 & 62.4 & 62.5 & 72.5 & 70.6 & 62.8 \\
DePT-G &53.7 & 55.7 & 55.8 & 58.2 & 56.0 & 61.8 & 57.1 & 69.2 & 66.6 & 72.2 & 76.3 & 63.2 & 67.9 & 71.8 & 68.2 & 63.6\\
SAR & 58.0 & 59.2 & 59.0 & 58.0 & 54.7 & 61.2 & 57.9 & 66.1 & 64.4 & 68.6 & 78.7 & 62.4 & 62.9 & 72.5 & 70.5 & 63.6 \\ 
\revise{ROID} & \revise{58.5} & \revise{59.8} & \revise{59.2} & \revise{57.6} & \revise{56.2} & \revise{61.4} & \revise{58.3} & \revise{67.5} & \revise{66.3} & \revise{70.3} & \revise{79.4} & \revise{62.1} & \revise{66.0} & \revise{73.5} & \revise{71.4} & \revise{64.5} \\

\rowcolor{gray!20}
\textbf{Ours} & \textbf{59.4} & \textbf{61.4} & \textbf{60.9} & \textbf{60.2} & \textbf{61.1} & \textbf{65.5} & \textbf{64.8} & \textbf{70.7} & \textbf{69.2} & \textbf{73.4} & 79.6 & \textbf{65.9} & \textbf{72.0} & \textbf{75.2} & \textbf{72.7} & \textbf{67.5} \\ 
\rowcolor{gray!20}
& ${\pm0.0}$ & ${\pm0.1}$ & ${\pm0.1}$ & ${\pm0.2}$ & ${\pm0.1}$ & ${\pm0.2}$ & ${\pm0.2}$ & ${\pm0.0}$ & ${\pm0.1}$ & ${\pm0.1}$ & ${\pm0.0}$ & ${\pm0.1}$ & ${\pm0.1}$ & ${\pm0.0}$ & ${\pm0.0}$ & ${\pm0.0}$ \\ 
\bottomrule
\end{tabular}
}
\end{center}
\end{table*}

\subsection{Implementation Details} 
In our experiments, all methods under various experimental conditions use the same architecture and pre-trained model parameters to ensure fair performance comparisons. We mainly use ViT-B/16 as the backbone for all experiments. In our experiments, the Factor Generator Network is a fully connected layer with input dimension \textit{d} and output dimension \textit{2d}. In addition, the Domain Separation Network is also a linear layer with input dimension \textit{d} and output dimension \textit{d}, which is used to generate Domain tokens. The combined additional parameters from these two networks increase the model's parameter count by approximately 1.7M. 
The learning rates for the Domain Separation Network and the Factor Generator Network are set to $0.2$ and $0.0005$, respectively. 
During training, we use the stochastic gradient descent (SGD) optimizer. The batch size used in the experiments is 64. During data loading, unlike SAR, we change the normalization settings to match the pre-trained timm model with mean = [0.5, 0.5, 0.5] and std = [0.5, 0.5, 0.5]. Our experimental results are reported as the mean and standard deviation obtained from three runs, each using a different randomly selected seed from the set \{2022, 2023, 2024\}. All experiments are tested on NVIDIA RTX3090. The pseudo-code of our proposed Progressive Conditioned Scale-Shift Recalibration (PCSR) method as shown in Algorithm \ref{alg: algorithm}.

\begin{algorithm}[!htbp]
\caption{Pseudo code of the proposed algorithm.}
\label{alg: algorithm}
\KwIn {Source pre-trained model $f_{\theta_s}$; target dataset $\{X_t\}$.}
\KwOut {The prediction of target samples $\{\hat{\mathbf{y}}\}$.}
Initialize the testing model $f_{\theta_t}$ with the source pre-trained model $f_{\theta_s}$ parameter weights; add all parameters $\tilde{\theta}$ to be fine-tuned to the optimizer; learning rate $\eta > 0$; \\
\For {batch $\mathbf{B}_j$ \textbf{in} $\{X_t\}$}{
 Compute the gradient $\nabla g$ with the loss function in equation \ref{eq:loss}\;
 Update $\tilde{\theta} \leftarrow \tilde{\theta}-\eta \nabla g$\;
 Output $\hat{\mathbf{y}}$ =  $f_{\theta_t}(\mathbf{B}_j)$\;
 }
\end{algorithm}

\begin{table}[!ht]
    \centering
    \caption{Classification Accuracy (\%)  for test-time adaptation from ImageNet to \textbf{ImageNet-R}, \textbf{ImageNet-A}, and \textbf{VisDA-2021} datasets.}
    \label{tab: imagenet-r}
    {
    \begin{tabular}{l|ccc}
    \toprule
        Method &  ImageNet-R &  ImageNet-A & VisDA-2021\\
    \midrule
        Source & 57.2 &31.1  &  57.7\\
        TENT & 61.3 &44.5 &  60.1\\
        SAR & 62.0 &45.3  & 60.1 \\
        \rowcolor{gray!20}
        \textbf{Ours} &   \textbf{66.5} &\textbf{52.1} & \textbf{64.8} \\
        \rowcolor{gray!20}
        &${\pm0.2}$& ${\pm0.3}$ &${\pm0.3}$ \\
    \bottomrule
    \end{tabular}   }
\end{table}

\subsection{Performance Results} 
\revise{
This section presents a comprehensive performance evaluation of our method across several benchmarks. We begin with detailed results on ImageNet-C, then evaluate cross-dataset generalization on ImageNet-R, ImageNet-A, and VisDA-2021, and finally assess scalability on the ViT-L/16 backbone. Additional experiments and ablations are included in the Appendix.
}
\subsubsection{Results on ImageNet-C}
We conducted a comprehensive evaluation of our method’s performance on the \textbf{ImageNet-C} dataset, with all data rounded to one decimal place to ensure accuracy and comparability. We report the mean and standard deviation values for each domain in the tables, based on three runs with different random seeds. Specifically, as shown in Table \ref{table: normal}, our method was rigorously tested across a diverse range of corruption types at the highest severity level (Level 5). The results demonstrate that across 15 different corruption scenarios, including noise, blur, weather, and digital corruption, our approach consistently outperforms others. Our method achieved an overall average accuracy of 67.5\%, which is 3.9\% higher than the second-best method, indicating its robustness and effectiveness in handling various challenging conditions. This substantial improvement highlights the strength of our approach in maintaining high performance, even in the presence of significant image distortions.

\subsubsection{Generalization to Other Datasets}
Furthermore, as shown in Table \ref{tab: imagenet-r}, our method also achieved a significant accuracy improvement on the \textbf{ImageNet-R} dataset, reaching 66.5\%, which is 5.2\% and 4.5\% higher than the previous methods, TENT and SAR, respectively. This demonstrates that our method exhibits strong robustness in handling domain adaptation issues within the ImageNet-R dataset. On the \textbf{ImageNet-A} dataset, our method also performed excellently, achieving an accuracy of 52.1\%, significantly surpassing TENT and SAR, which achieved an accuracy of 44.5\% and 45.3\%, respectively. This improvement highlights the strong generalization ability of our method in handling challenging adversarial selected image scenarios. On the \textbf{VisDA-2021} dataset, our method achieved an accuracy of 64.8\%. This result outperforms the baseline methods, TENT and SAR, which both achieved an accuracy of 60.1\%. The consistent improvements across multiple datasets further underscore the effectiveness of our method in addressing various domain shifts.

\subsubsection{Scalability Analysis on ViT-L}
\revise{We extend our experimentation to a larger ViT-L/16 backbone. 
The learning rates of the Factor Generator Network and the Domain Separation Network are set 
to $0.0001$ and $0.05$ respectively. The results, as illustrated in Table \ref{tab: vit-l-normal}, showcase the superiority of our proposed PCSR method over the baseline SAR. This robust performance demonstrates the efficiency of our proposed PCSR method across diverse transformer backbones.}

\begin{table*}[!t]
    \centering
    \caption{Classification Accuracy (\%)  in \textbf{ImageNet-C} with \textbf{ViT-L/16} at the highest severity (Level 5).}
    \label{tab: vit-l-normal}
    \resizebox{\linewidth}{!}{
\begin{tabular}{l|ccccccccccccccc|c}
\toprule
Method & gaus & shot & impul & defcs & gls & mtn & zm & snw & frst & fg & brt & cnt & els & px & jpg & Avg.\\
    \midrule
        Source & 62.1 & 61.4 & 62.3 & 52.7 & 45.1 & 60.6 & 55.1 & 66.2 & 62.4 & 62.5 & 80.2 & 39.8 & 56.2 & 74.3 & 72.7 & 60.9 \\
        TENT & 65.6 & 68.3 & 67.6 & 63.4 & 59.9 & 66.8 & 60.7 & 69.0 & 68.5 & 67.4 & 81.0 & 28.9 & 64.7 & 77.2 & 74.7 & 65.6 \\
        SAR & 66.0 & 66.6 & 66.2 & 61.3 & 55.1 & 66.1 & 58.3 & 68.4 & 65.7 & 66.3 & 81.0 & 26.8 & 63.7 & 74.5 & 73.6 & 64.0 \\
        \rowcolor{gray!20}
        \textbf{Ours} & \textbf{68.5} & \textbf{69.1} & \textbf{68.9} & \textbf{65.5} & \textbf{64.6} & \textbf{68.7} & \textbf{67.4} & \textbf{72.7} & \textbf{70.3} & \textbf{69.6} & \textbf{81.9} & \textbf{63.5} & \textbf{71.6} & \textbf{77.6} & \textbf{76.3} & \textbf{70.4} \\
        \rowcolor{gray!20}
        & ${\pm0.1}$ & ${\pm0.5}$ & ${\pm0.1}$ & ${\pm0.4}$ & ${\pm0.1}$ & ${\pm0.4}$ & ${\pm0.1}$ & ${\pm0.1}$ & ${\pm0.3}$ & ${\pm2.3}$ & ${\pm0.1}$ & ${\pm1.1}$ & ${\pm0.0}$ & ${\pm1.0}$ & ${\pm0.0}$ & ${\pm0.3}$ \\
    \bottomrule
    \end{tabular}   }
\end{table*}

\begin{table}[!ht]
    \centering
    \caption{Ablation study about the condition with the Class token and the Domain token at the highest severity of ImageNet-C dataset.}
\label{table: ablation}
{
\begin{tabular}{l|cc}
\toprule
Method  & Avg. \\	
\midrule
Baseline  Method & 63.6 \\
\quad\quad  + Conditioned on Class token  & 66.7\\
\quad\quad + \revise{Conditioned on Mean token}  & \revise{66.8}\\
\quad\quad  + Conditioned on Domain token   & 67.1 \\
\rowcolor{gray!20}
\textbf{Our Method}  & \textbf{67.5}\\
\bottomrule
\end{tabular}
}
\end{table}

\subsection{Ablation Studies and Design Analysis}
\subsubsection{Effect of Conditioning Tokens}
We conducted an ablation study to analyze the individual effects of using the Class token and Domain token to generate the scale and shift factors in our model. As shown in Table 3, the results reveal the impact of each component on the model's performance.
It can be seen that the baseline method achieved an average accuracy of 63.6\%. When the Class token was introduced for calculating the scale and shift factors, the accuracy increased to 66.7\%, indicating that the Class token significantly enhances the model's ability to adapt to domain shifts. \revise{When using only the mean token, the accuracy reached $66.8\%$, whereas using only the domain token generated by the Domain Separation Network to calculate the scale and shift factors improved the accuracy to $67.1\%$, demonstrating its strong individual contribution.}
 Finally, the method that integrated both the Class token and Domain token for calculating the scale and shift factors achieved the highest accuracy of 67.5\%. This result highlights the complementary nature of these two components, leading to the best overall performance.

\revise{
\begin{table}[htbp]
\centering
\caption{Comparison of feature aggregation methods after DSN.}
\label{tab:aggregation}
\begin{tabular}{lcc}
\toprule
\textbf{Aggregation Method} & \textbf{Accuracy (\%)} & \textbf{Extra Parameters} \\
\midrule
Max Pooling       & 66.9 & No \\
Average Pooling   & 67.5 & No \\
Attention-based   & 67.5 & Yes \\
\bottomrule
\end{tabular}
\end{table}
}

\revise{\subsubsection{Aggregation method comparison after Domain Separation Network} 

To identify an effective feature aggregation strategy following the Domain Separation Network (DSN), we conducted experiments comparing three approaches: max pooling, average pooling, and attention-based aggregation. The experimental results are summarized in Table~\ref{tab:aggregation}, where both average pooling and attention-based aggregation achieve an accuracy of 67.5\%, indicating only a marginal difference. However, the attention-based method introduces additional trainable parameters, whereas max pooling yields a lower accuracy of 63.7\%. Considering both performance and computational efficiency, we ultimately adopt the average pooling strategy.}

\revise{
\begin{table}[!ht]
\centering
\caption{Comparison of individual and shared adaptation of Q, K, V in terms of accuracy and training time.}
\label{tab:qkv_param_recali}
\begin{tabular}{l|cc}
\toprule
\textbf{Metric} & \textbf{Accuracy (\%)} & \textbf{Running Time (min)} \\
\midrule
Independent & 67.4 & $\sim 18$ \\
shared & 67.5 & $\sim 15$ \\
\bottomrule
\end{tabular}
\end{table}
}

\revise{
\subsubsection{Analysis of QKV Parameter Sharing} 
We evaluated whether independent scale and shift parameters for Q, K, and V are necessary by comparing the individual adaptation with the shared adaptation (QKV) under identical hyperparameters. As shown in Table~\ref{tab:qkv_param_recali}, the difference in accuracy is negligible (67.5\% vs 67.4\%), while the shared parameter design slightly decreases the training time ($\sim 18$ min vs. $\sim 15$ min) but reduces memory usage. Therefore, we adopt shared adaptation to simplify the model, achieving comparable performance while saving memory and computational resources.
}

\revise{
\begin{figure}[htbp]
    \centering
    \includegraphics[width=1\linewidth]{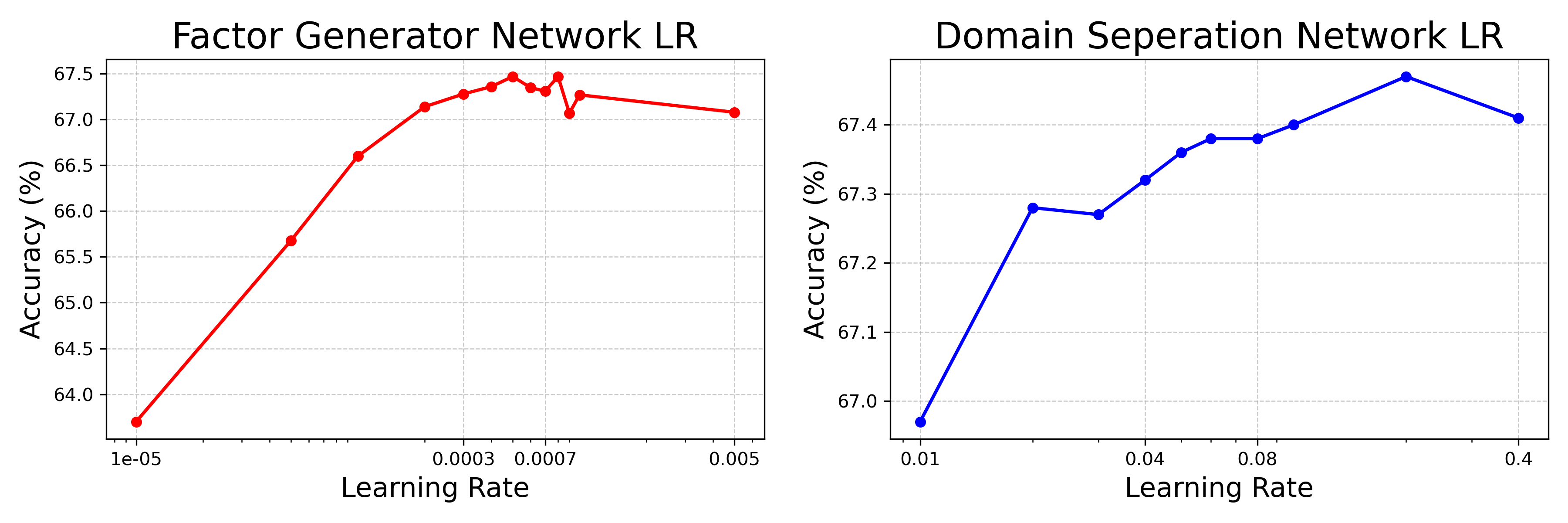}
    \caption{Learning Rate Sensitivity Analysis}
    \label{fig:lr_all}
\end{figure}
}
\revise{\subsubsection{Learning Rate Sensitivity}
Figure~\ref{fig:lr_all} presents the sensitivity of both networks to different learning rates. The Factor Generator Network achieves optimal accuracy with a smaller learning rate of $0.0005$, while the Domain Separation Network performs best with a larger rate of $0.2$. These values are therefore adopted in our subsequent experiments.}

\revise{
\section{Future work}
In the current study, our experiments are conducted primarily on image classification tasks using Transformer-based architectures, through which we have verified the effectiveness of our approach. For future work, we intend to explore the effectiveness and generalizability of our methods across a broader range of model architectures, including CNN-based networks, and to further extend their applicability to more challenging tasks such as object detection and semantic segmentation.
}

\section{Conclusion}
\label{sec:conclusion}
In this work, we addressed the critical challenge of online test-time adaptation in Vision Transformers, focusing on mitigating performance degradation caused by domain shifts. Our proposed approach, Progressive Conditioned Scale-Shift Recalibration (PCSR), introduces a novel structure to dynamically adjust the self-attention modules during inference. By incorporating scale and shift factors into the Query, Key, and Value components, and conditioning these factors on a domain-specific Domain token, our method effectively adapts the model to new domains in real-time. We develop a Domain Separation Network to generate a Domain token, which represents the domain shift derived from all the patch tokens at each transformer network layer. Once the Domain token is learned, we train a Factor Generator Network to produce scale and shift factors conditioned on the Domain token and the Class token. These factors, applied at each transformer network layer, gradually mitigate the impact of domain shifts during online test-time adaptation. Experimental results on benchmark datasets demonstrate that the proposed Progressive Conditioned Scale-Shift Recalibration method significantly enhances online test-time domain adaptation performance.

\section{Acknowledgment}
This work was supported by the National Natural Science Foundation of China (No. 62331014) and Project 2021JC02X103.

\bibliographystyle{IEEEtran}
\bibliography{IEEEabrv,ref}

\begin{IEEEbiography}[{\includegraphics[width=1in,height=1.25in,clip,keepaspectratio]{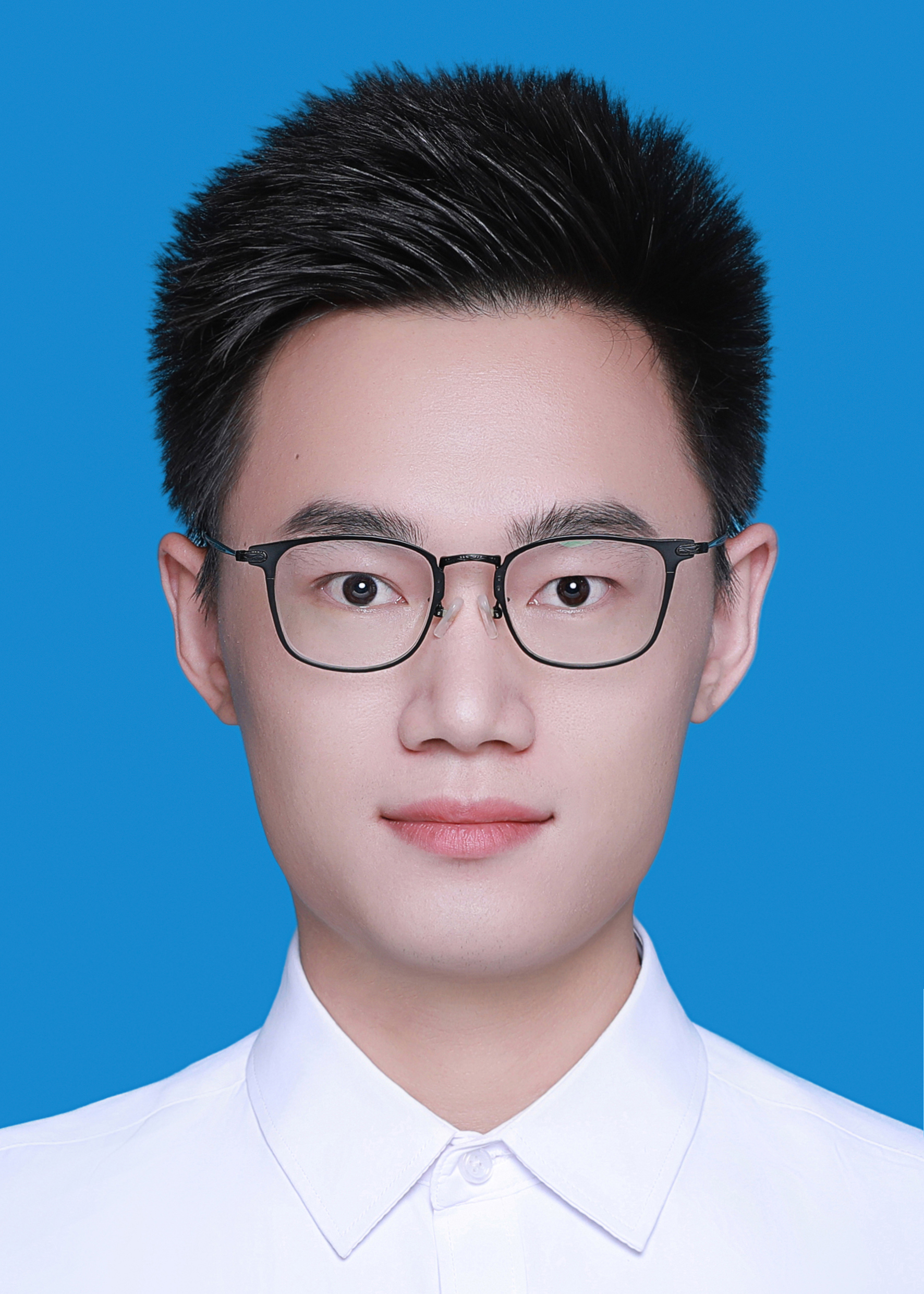}}]{Yushun Tang}
is currently pursuing a Ph.D. in Intelligent Manufacturing and Robotics at the Southern University of Science and Technology (SUSTech). He holds a Bachelor’s degree in Optoelectronic Information Science and Engineering from Harbin Engineering University and a Master’s degree in Electronic Science and Technology from SUSTech. His current research focuses on Computer Vision, Transfer Learning, Domain Adaptation, and Text-to-Image Generation.
\end{IEEEbiography}

\begin{IEEEbiography}
[{\includegraphics[width=1in,height=1.25in,clip,keepaspectratio]{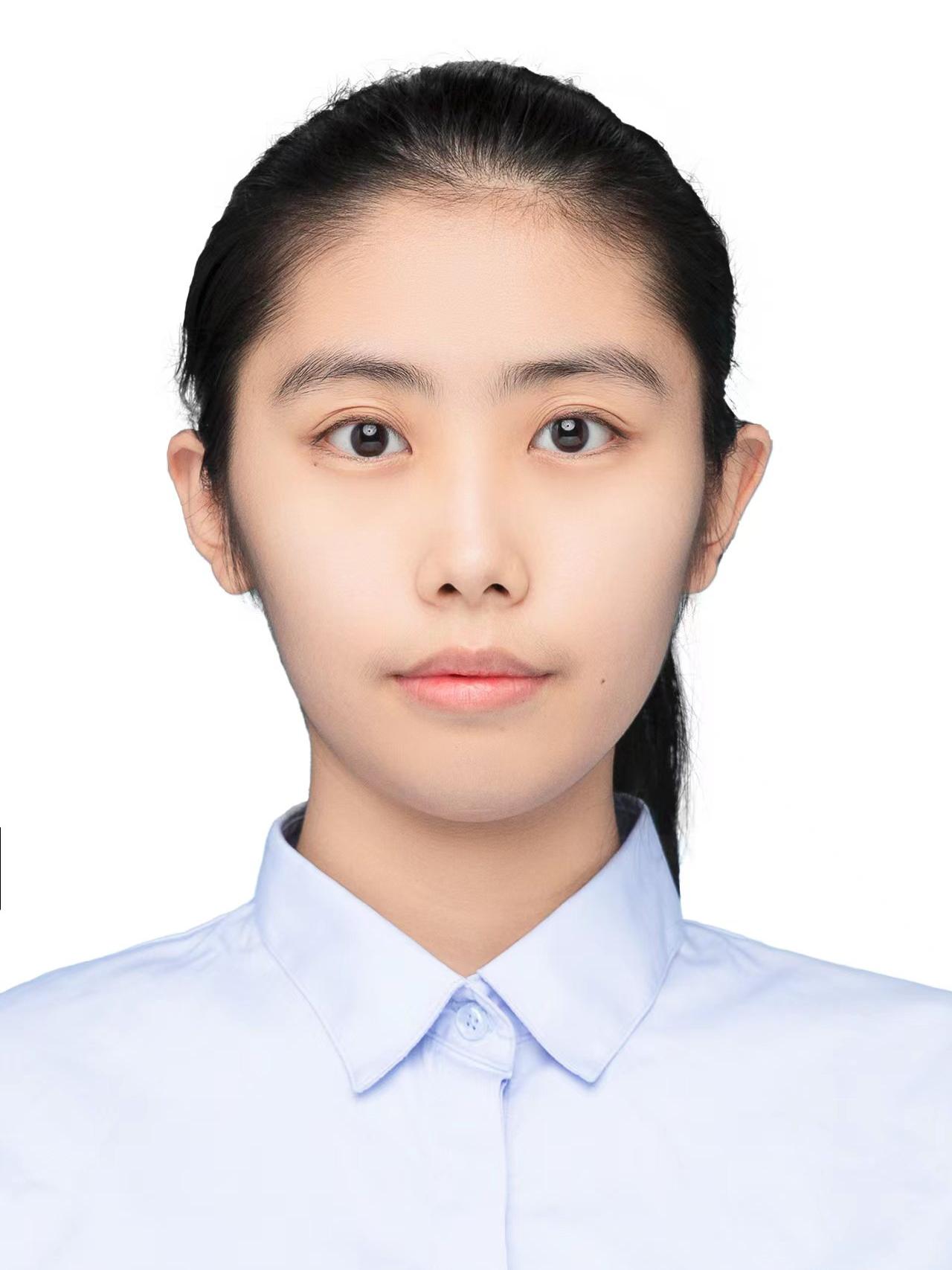}}]{Ziqiong Liu}
is currently pursuing a Master’s degree in Electronic Science and Technology at the Southern University of Science and Technology (SUSTech). She holds a Bachelor’s degree in Information Engineering from SUSTech. Her current research focuses on Computer Vision, Transfer Learning, and Domain Adaptation.
\end{IEEEbiography} 

\begin{IEEEbiography}
[{\includegraphics[width=1in,height=1.25in,clip,keepaspectratio]{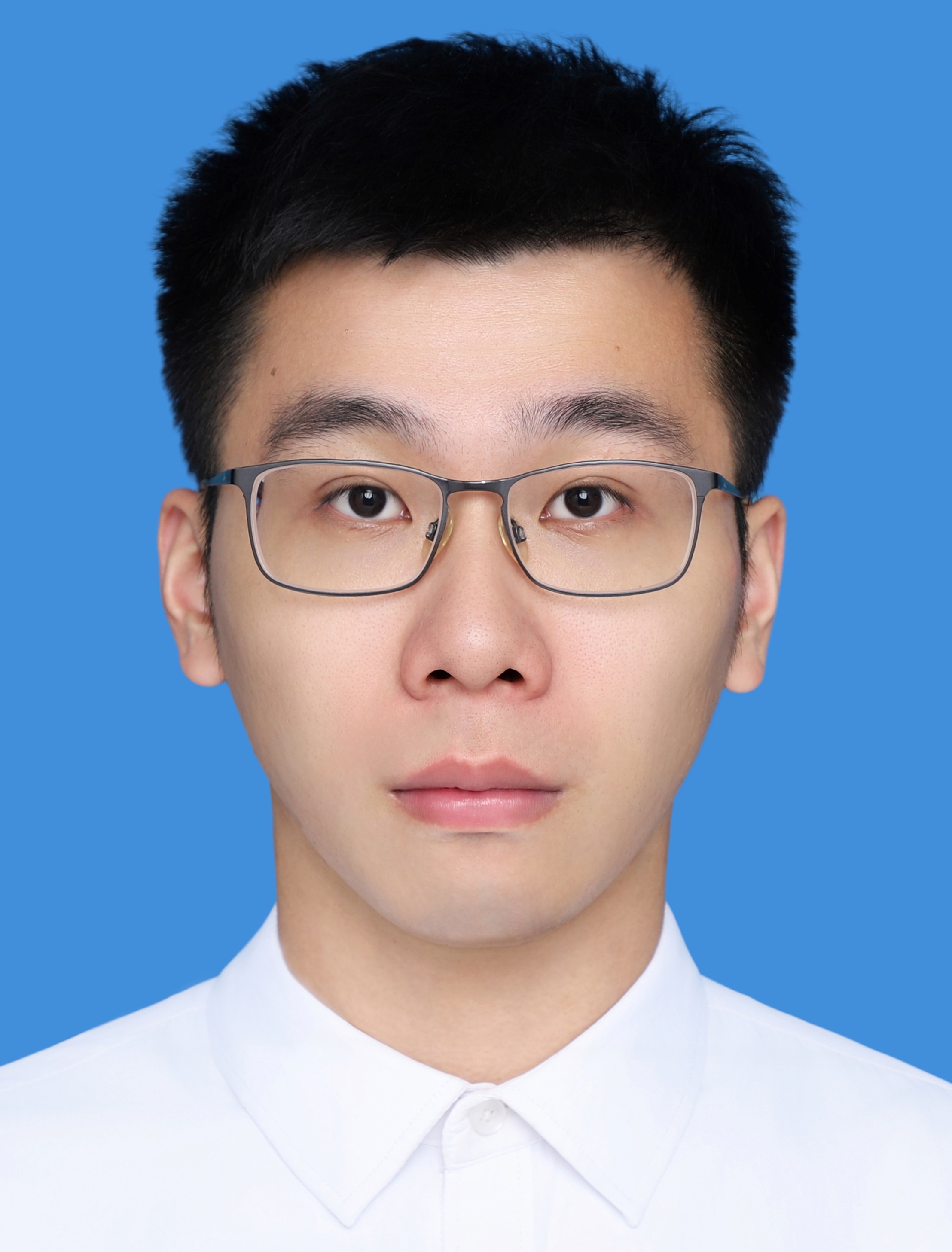}}]{Jiyuan Jia}
is currently pursuing a Master’s degree in Electronic Science and Technology at the Southern University of Science and Technology (SUSTech). He holds a Bachelor’s degree in Robotics Engineering from SUSTech. Her current research focuses on Computer Vision.
\end{IEEEbiography}

\begin{IEEEbiography}
[{\includegraphics[width=1in,height=1.25in,clip,keepaspectratio]{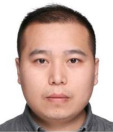}}]{Yi Zhang}
is currently pursuing a Ph.D. in Artificial Intelligence at the Southern University of Science and Technology (SUSTech) and Harbin Institute of Technology, with an expected graduation date in July 2025. Prior to this, he earned a Bachelor’s degree in Software Engineering from Northeastern University (China) and a Master’s degree in Information Systems from The University of Texas. His current research focuses on vision-language models, Few-shot Learning, and Visual Reasoning.
\end{IEEEbiography}

\begin{IEEEbiography}
[{\includegraphics[width=1in,height=1.25in,clip,keepaspectratio]{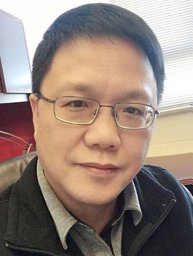}}]{Zhihai He} (Fellow, IEEE) received the B.S. degree in mathematics from Beijing Normal University, Beijing, China, in 1994, the M.S. degree in mathematics from the Institute of Computational Mathematics, Chinese Academy of Sciences, Beijing, China, in 1997, and the Ph.D. degree in electrical engineering from the University of California, at Santa Barbara, CA, USA, in 2001. In 2001, he joined Sarnoff Corporation, Princeton, NJ, USA, as a member of technical staff. In 2003, he joined the Department of Electrical and Computer Engineering, University of Missouri, Columbia, MO, USA, where he was a tenured full professor. He is currently a chair professor with the Department of Electrical and Electronic Engineering, Southern University of Science and Technology, Shenzhen, P. R. China. His current research interests include image/video processing and compression, wireless sensor network, computer vision, and cyber-physical systems.

He is a member of the Visual Signal Processing and Communication Technical Committee of the IEEE Circuits and Systems Society. He serves as a technical program committee member or a session chair of a number of international conferences. He was a recipient of the 2002 {\sc IEEE Transactions on Circuits and Systems for Video Technology} Best Paper Award and the SPIE VCIP Young Investigator Award in 2004. He was the co-chair of the 2007 International Symposium on Multimedia Over Wireless in Hawaii. He has served as an Associate Editor for the {\sc IEEE Transactions on Circuits and Systems for Video Technology} (TCSVT), the {\sc IEEE Transactions on Multimedia} (TMM), and the Journal of Visual Communication and Image Representation. He was also the Guest Editor for the IEEE TCSVT Special Issue on Video Surveillance.

\end{IEEEbiography}

\clearpage
\appendices
\section{Supplementary Experimental Results}


\subsection{Results on ImageNet-C at Severity Level 3}

\begin{table*}[!t]
\centering
\caption{Classification accuracy (\%) for each corruption in \textbf{ImageNet-C} at severity level 3. The best results are shown in \textbf{bold}.}
\label{table: normal_level_3}
\resizebox{\linewidth}{!}{
\begin{tabular}{l|ccccccccccccccc|c}
\toprule
Method & gaus & shot & impul & defcs & gls & mtn & zm & snw & frst & fg & brt & cnt & els & px & jpg & Avg. \\	
\midrule
Source & 72.1 & 71.5 & 71.3 & 62.3 & 51.1 & 69.1 & 59.1 & 69.0 & 60.0 & 70.1 & 80.1 & 74.0 & 75.1 & 77.6 & 75.2 & 69.2 \\
TENT   & 74.3 & 73.9 & 73.6 & 70.8 & 66.6 & 73.7 & 66.9 & 73.2 & 68.7 & 76.0 & 81.6 & 78.9 & 78.5 & 79.7 & 77.1 & 74.2 \\
SAR    & 74.3 & 73.9 & 73.7 & 70.9 & 66.5 & 73.8 & 66.9 & 73.1 & 68.7 & 75.8 & 81.8 & 78.9 & 78.5 & 79.8 & 77.1 & 74.2 \\ 
\rowcolor{gray!20}
\textbf{Ours} & \textbf{74.9} & \textbf{74.7} & \textbf{74.5} & \textbf{72.2} & \textbf{70.9} & \textbf{75.5} & \textbf{71.2} & \textbf{75.4} & \textbf{72.3} & \textbf{78.0} & \textbf{81.9} & \textbf{79.8} & \textbf{79.9} & \textbf{80.4} & \textbf{78.9} & \textbf{76.0} \\ 
\rowcolor{gray!20}
& ${\pm0.0}$ & ${\pm0.0}$ & ${\pm0.1}$ & ${\pm0.1}$ & ${\pm0.0}$ & ${\pm0.0}$ & ${\pm0.1}$ & ${\pm0.0}$ & ${\pm0.0}$ & ${\pm0.0}$ & ${\pm0.0}$ & ${\pm0.1}$ & ${\pm0.0}$ & ${\pm0.0}$ & ${\pm0.1}$ & ${\pm0.0}$ \\
\bottomrule
\end{tabular}
}
\end{table*}

In Table~\ref{table: normal_level_3}, we report the classification accuracy (\%) of different test-time adaptation methods on ImageNet-C at corruption severity level 3. The experiments follow the same methodology and parameters as the level-5 evaluation, with the only change being the use of level-3 test data. Our proposed PCSR method achieves the highest average accuracy of \textbf{76.0\%}, consistently outperforming TENT (74.2\%) and SAR (74.2\%) across all 15 corruption types, delivering performance gains of up to 2.2\% in individual categories while maintaining stable improvements overall.

\subsection{Results under Label and Mixed Distribution Shifts}

\begin{table}[htbp]
\centering
\caption{Comparison under different real-world scenarios: L.S. denotes label distribution shifts, and M.S. denotes mixed domain shifts.}
\label{tab:label_and_mix_shifts}
\begin{tabular}{lcc}
\toprule
\textbf{Method} & \textbf{L.S} & \textbf{M.S} \\
\midrule
Source & 29.9 & 29.9 \\
TENT   & 47.3 & 16.5 \\
SAR    & 58.0 & 57.7 \\
OURS   & 63.0 & 61.2 \\
\bottomrule
\end{tabular}
\end{table}

Table~\ref{tab:label_and_mix_shifts} presents the adaptation performance on ImageNet-C (severity level 5) under two types of distribution shifts: label shifts (L.S.) and mixed domain shifts (M.S.). Compared with existing methods, our approach achieves consistently higher accuracy in both scenarios, demonstrating superior robustness and adaptability to diverse real-world distribution changes.

\end{document}